\documentclass[10pt,journal,compsoc]{IEEEtran}

\usepackage{times}  %Required
\usepackage{helvet}  %Required
\usepackage{courier}  %Required

\usepackage{url}
\usepackage{graphicx}
\usepackage{color}
\usepackage{xspace}

\usepackage[normalem]{ulem}
\usepackage{dsfont}
\usepackage[caption=false,font=footnotesize]{subfig}
\usepackage{makeidx}  % allows for indexgeneration
\usepackage{multirow}
\usepackage{amsmath,amsfonts,amssymb,amsthm}
\usepackage{breqn}

\usepackage{hyperref}
\usepackage{cleveref}

\newcommand{\bd}[1]{\boldsymbol{#1}}

\newcommand{\s}[1]{\mathcal{#1}}
\newcommand{\f}[1]{\mathds{#1}}

\DeclareMathOperator*{\argmax}{arg\,max}

\usepackage{ulem}
\usepackage{cancel}

\newcommand\redsout{\bgroup\markoverwith
{\textcolor{red}{\rule[0.5ex]{2pt}{0.4pt}}}\ULon}

\definecolor{berdeiluna}{rgb}{0.2,0.6,0.2}

\definecolor{sarandonga}{rgb}{0.65,0.35,0}

\definecolor{urdin}{rgb}{0.4,0,0.6}

\begin{document}

\title{Candidate labeling for crowd learning}

\author{Iker~Be{\~n}aran-Mu{\~n}oz,
        Jer{\'o}nimo~Hern{\'a}ndez-Gonz{\'a}lez,
        and~Aritz~P{\'e}rez%
\IEEEcompsocitemizethanks{\IEEEcompsocthanksitem I. Be{\~n}aran-Mu{\~n}oz and A. P{\'e}rez are with the Basque Center for Applied Mathematics, Bilbao, Spain.
\{ibenaran, aperez\}@bcamath.org
\IEEEcompsocthanksitem Jer{\'o}nimo~Hern{\'a}ndez-Gonz{\'a}lez is with the %Department of Computer Science and Artificial Intelligence, 
University of the Basque Country UPV/EHU, Donostia, Spain. %\protect\\
% note need leading \protect in front of \\ to get a newline within \thanks as
% \\ is fragile and will error, could use \hfil\break instead.
E-mail: jeronimo.hernandez@ehu.eus
}% <-this % stops an unwanted space
\thanks{}}

% \received{1 May 2013}
% \finalform{10 May 2013}
% \accepted{13 May 2013}
% \availableonline{15 May 2013}
% \communicated{S. Sarkar}

\maketitle

\begin{abstract}
Crowdsourcing has become very popular among the machine learning community as a way to obtain labels that allow a ground truth to be estimated for a given dataset. 
In most of the approaches that use crowdsourced labels, annotators are asked to provide, for each presented instance, a single class label. %Most of the approaches that use crowdsourced labeled data ask the labelers to provide a single class label for each presented instance. 
Such a request could be inefficient: considering that the labelers may not be experts, proceeding in this way could fail to take real advantage of the knowledge of the labelers. In this paper, the use of candidate labeling for crowd learning is proposed, where the annotators may provide more than a single label per instance to try not to miss the real label. %make sure that the real label is not missed. 
The main hypothesis is that, by allowing candidate labeling, knowledge can be extracted from the labelers more efficiently than in the standard crowd learning scenario. Empirical evidence which supports that hypothesis is presented.
\end{abstract}


%\linenumbers

%% main text

\section{Introduction}\label{sec:introduction}

In the last few decades, the web and the community behind it have become a huge source of data. Among different ways of taking advantage of it, \emph{crowdsourcing} has emerged as a popular strategy to get things done in a collaborative way thanks to the involvement of a crowd of \emph{workers}. Commonly, in exchange for a certain type of incentive, the workers complete short and simple tasks. 
%, usually called \emph{human intelligence tasks}. 
Crowdsourcing has been widely used to solve different kinds of problems, such as text correction \cite{bernstein15}, text translation \cite{corney10} or malaria diagnostics \cite{luengo12}, and platforms such as Amazon Mechanical Turk or CrowdFlower have boosted its popularity. 
%The machine learning community has seen in this strategy an unparalleled opportunity to label datasets which otherwise would remain unsupervised. The task of labeling, which is key in standard supervised classification, cannot always be carried out in the traditional way due to its complexity and/or cost. 
In machine learning, the task of labeling, which is key in standard supervised learning, cannot always be carried out in a traditional way due to its complexity and/or cost. Crowdsourcing is an unparalleled opportunity to get datasets labeled.
In the absence of the real fully reliable labeling of a dataset (a.k.a. ground truth), crowd labeling consists of obtaining the labels of the training examples from a crowd of workers, a.k.a. annotators or labelers in this context \cite{snow08,sheng08}.  As the reliability of the annotators cannot be guaranteed, a number of them are usually asked to label the same example in the hope that the consensus label is the correct one.

Subsequently, the problem of learning from a dataset labeled by a crowd is a challenge in itself. 
%Known in the related literature as \emph{crowd learning} or \emph{learning from crowds} \citep{raykar10},
In \emph{crowd learning}~\cite{raykar10}, 
the objective is to estimate a realistic ground truth from examples labeled by multiple annotators in order to learn a classification model. If the collected labels fulfill certain conditions, crowd learning can be as reliable as learning from a single expert in a traditional way \cite{snow08,sheng08}. 
Depending on the domain, it might be even more efficient, in terms of time and cost, than the traditional approach. 
The EM-based proposal by \cite{dawid79}, which, for each worker, estimates the probability of confusing two class labels in order to estimate a ground truth, and learns a model simultaneously, has become a standard in the field. Many posterior frameworks are based on it \cite{raykar10,wang11}. 
Although some methods \cite{jin02} disregard the individual information about the annotators and aggregate the labels to learn with a probability distribution over the class labels per example, the expertise of each individual worker is usually inferred \cite{demartini12,whitehill09,hernandez15,raykar10,welinder10}. 
%There are some methods \citep{jin02} which disregard the individual information about the annotators and aggregate the labels to learn with a probability distribution over the class labels for each example. However, it is more common to try to infer the expertise of each individual worker \citep{demartini12,whitehill09,hernandez15,raykar10,welinder10}. 
Instance difficulty has also been modeled \cite{whitehill09}, as well as worker competence and the bias of the annotations \cite{welinder10}.

Thus, most of the effort is in modeling and dealing with the (un)reliability of the annotators, 
as it is generally accepted that annotators may have limited knowledge about the assigned task. However, they are asked to provide a single label ---the preferred one--- given an (incomplete) instance. These two ideas may seem contradictory: being strict with someone who might not be able to help us. This traditional approach is referred to as \textbf{full labeling} throughout this work. In this paper, the central idea is that a more relaxed request could allow for the extraction of more knowledge from the available annotators. For example, if an annotator is in doubt between two or more labels and they are forced to choose only one, they may pick the wrong one. On the contrary, if they are allowed to provide more than one label per instance, following the same example, the worker could select both. In this way, one might expect a lower number of mistakes, that is, they will include the correct class label in their sets of selected labels with high probability. Moreover, knowing that an annotator doubts between a few class labels provides useful information about the underlying distribution of labels and the ground truth.

Frameworks where the single-label request is relaxed have already been proposed, such as the works by \cite{grady10} and \cite{smyth94}, where annotators can say how sure they are about their annotations, or other works where annotators are allowed to claim that they do not know the answer \cite{zhong15,venanzi16}. Our proposal, inspired by the subfield of weak supervision \cite{hernandez16}, may be seen as a step forward in this direction. Weakly supervised problems are characterized by the lack of a full labeling. One of them is the partial or candidate labels \cite{cour11} problem, which assumes that all the training examples are provided together with a set of labels, with the guarantee that the real label is in that set. This concept is extended to the context of crowd learning and allows annotators to provide as many labels as they want when they are not able to choose a single one. A similar problem has been studied under the name of \emph{approval voting}~\cite{brams78,falmagne96,procaccia15}. However, the research performed in this field is not of our interest since the aim is not to infer a ground truth or of learn any model. Moreover, it has been applied in social sciences rather than in a machine learning context.

The main intuition behind this study is that learning from crowds can be more efficient in terms of time and number of annotators when annotators are allowed to provide this kind of labeling. In \cite{banerjee17}, they already provided some evidence that workers answer faster using candidate labeling (\emph{checkbox interface} according to their terminology) than using the traditional labeling (\emph{radio button interface}). The aim of this paper is to show that not only is this method less costly, but that more knowledge can be extracted and hence better results can be obtained. The main contribution of this work is the notion of candidate labeling applied to crowd learning, which offers a new framework for annotation gathering, along with candidate voting, for inferring the ground truth from it. An empirical study of the performance of this method is presented, comparing the results of the popular majority voting strategy for annotators that carry out a full labeling against a voting scheme adapted for examples labeled through candidate labeling.

The article is structured as follows: First, the problem is formally described. In Section~\ref{sec:experiments}, an explanation is given about the simulation of annotators to produce candidate labeling by modeling instance difficulty and annotator behavior. Next, the results of the experimentation are discussed. Finally, in Section~\ref{sec:conclusions} conclusions are drawn and possible future steps are discussed.

\section{Candidate labeling for crowd learning}
\label{sec:candidatecrowd}

\textbf{Crowd labeling} makes use of a set of \textbf{annotators}, $\s{L}$, which are asked to label a set of $n$ unlabeled instances. Each instance belongs to a class from the set of possible class labels $C$. In this work, we assume that each instance is labeled by a fixed number $l$ of different annotators. The generalization to scenarios where every instance is not labeled by the same number of annotators is straightforward. It is also assumed that $|C|=r>2$ and that there is a single \textbf{ground truth} label $c^*$ for each instance.

Once the labels given by the annotators have been collected, a single label can be assigned to each instance by using an aggregation function. One simple yet effective aggregation functions in the traditional full labeling framework is the (majority) \textbf{voting}, which will be referred to as \textbf{full voting}:
\begin{equation}
v(L) = \argmax_c\sum_{j=1}^{l} \mathds{1}(c_j=c)
\label{eq:voting}
\end{equation}
where $L=\{c_j\}_{j=1}^l$ is the multiset of labels provided by the annotators in $\s{L}$ and $\f{1}(a)$ is a function which returns 1 if the condition $a$ is $true$ and $0$ otherwise.

It is well known that, by using the voting function, high quality ground truth labels can be obtained even with unreliable labelers \cite{snow08,hernandez15} and, under mild conditions regarding the reliability of the annotators, its error tends to be smaller as the number of annotators $l$ increases. However, in practice, in order to estimate the real ground truth, the number of annotators $l$ cannot grow arbitrarily, and it is commonly limited by the available economical resources. We believe that it is possible to achieve a reduction of the error with a tighter budget, by allowing for a more flexible and relaxed labeling scheme.

Under the term of \textbf{weak supervision}, many generalizations of the supervised classification problem regarding the uncertainty surrounding the class variable are considered \cite{hernandez16}. In this work, we are particularly interested in the \textbf{partial} or \textbf{candidate labels} problem \cite{cour11}, in which 
%the training set consists of instances, each one of them provided with a set of labels $S\subseteq C$ instead of a single one. 
each training instance is provided with a set of labels $S\subseteq C$ instead of a single one. 
In that problem, it is assumed that the ground truth label $c^*$ belongs to $S$ and there is no restriction on the size of the sets of labels $S$, which can vary from one instance to another.
%
%Following the aforementioned candidate labels problem, 
In this work, the \textbf{candidate labeling} for crowd learning, an extension of the traditional full labeling, is proposed. In candidate labeling, given an unlabeled instance, each available annotator is asked to provide the set of most promising labels $S\subseteq C$ for it according to their knowledge. Each set of labels $S$ provided by a labeler is a \textbf{candidate set}. The inclusion of the ground truth label in the candidate set is a reasonable assumption, unlike in \cite{cour11}, where it is guaranteed. It is assumed that, depending on the difficulty of the instance and the behavior of the annotator, the size of $S$ could vary. In candidate labeling, for each instance, a set $L=\{S_j\}_{j=1}^l$ of candidate sets is obtained from the annotators in $\s{L}$. For this setting, the use of the \textbf{candidate voting} function is proposed for estimating the ground truth:
\begin{equation}
\nu(L) = \argmax_c\sum_{j=1}^{l} \frac{1}{|S_j|}\mathds{1}(c \in S_j)
\label{eq:cVoting}
\end{equation}
Candidate voting is a simple way for aggregating the labels provided by the annotators. It should be noted that it is a natural generalization of the full voting strategy in the candidate labeling context. In fact, this function behaves as majority voting (Equation \ref{eq:voting}) if all the annotators provide $S_j$ such that $|S_j|=1$. In practice, ties (i.e., when two or more class labels obtain the maximum number of votes) are solved randomly. The goal is to create a labeling that minimizes the (aggregation) \textbf{error} $\epsilon(\nu)=E[\f{1}(c^*\neq\nu(L))]$, where $\nu(L)$ is the aggregated label.

The main idea of this work is that candidate labeling is more efficient than full labeling regarding the extraction of knowledge from a reduced number of annotators $l$, especially when they doubt between several class labels. That is, given a sufficiently small number of annotators $l$, the error of the candidate labeling tends to be smaller or equal to the error of the full labeling. In this work, empirical evidence is provided in order to confirm the following closely related \textbf{hypotheses}:
  \begin{itemize}
      \item \textbf{H1}: candidate labeling requires (equal or) \emph{fewer annotators} than the full labeling to achieve (equal or) lower error,
      \item \textbf{H2}: the number of annotators required by full labeling to achieve the performance of candidate labeling grows as \emph{the difficulty of instances grows},
      \item \textbf{H3}: the difference in the error is higher as the \emph{number of possible class labels} increases,
      \item \textbf{H4}: the difference in the error is higher with \emph{more hesitant annotators}.
\end{itemize}

\section{Empirical evidence}
\label{sec:experiments}

\begin{figure*}
\centering
\includegraphics[width=0.99\textwidth]{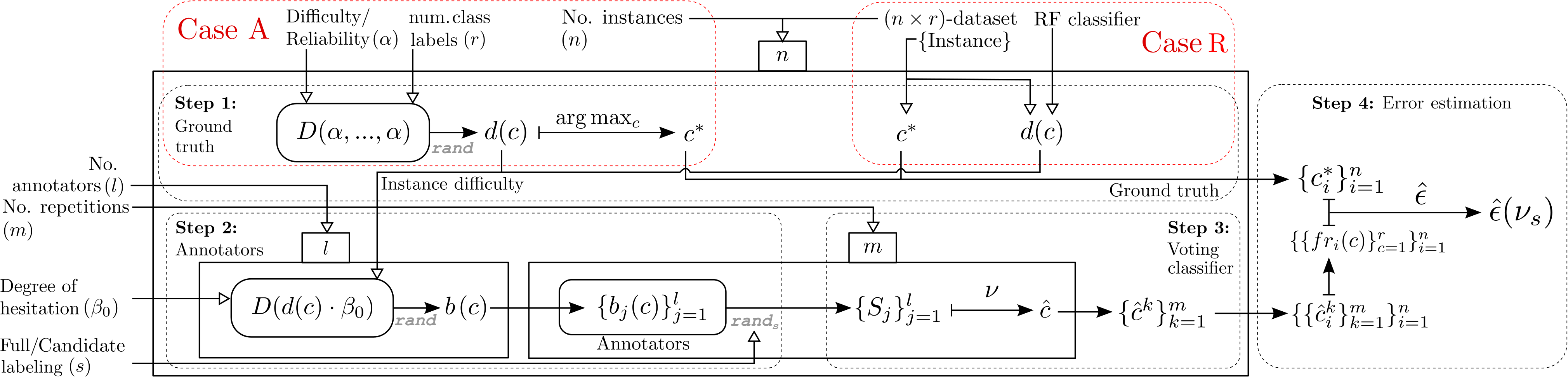}
\caption{Overview of the experimental framework with artificial domains.}
\label{fig:artDomains}
\end{figure*}

In this section, empirical evidence that supports the hypotheses \textbf{H1}, \textbf{H2}, \textbf{H3} and \textbf{H4} is provided. For this purpose, the error of full and candidate labeling will be estimated and analyzed under different experimental conditions, which involve a varying number of annotators (together with their degree of hesitation), difficulty of the instance, and number of available labels ($r$). In this work, we have decided not to study the reliability of the workers. This characteristic would definitely have an impact on the results, obscuring the contribution of the candidate labeling. In order to explicitly control all the synthetic conditions and to develop simple and intuitive scenarios, experiments are carried out on artificial domains. Two types of simulations are performed: In one case (referred to as Case~A hereinafter) the simulations are completely artificial, and, in the other one (Case~R), they are derived from real-world datasets.

\subsection{Data generation: Case~A}
\label{sec:data}

In Case~A, the difficulty of the instances and the ground truth are generated artificially. The experimental framework is summarized in Fig.~\ref{fig:artDomains}. It consists of four steps. %Case~A and Case~R only differ in the Step~1. As in the experiments with real data (Case~R) the ground truth labeling is available, this first step limits to measuring instance difficulty. In Case~A, the ground truth label as well as the instance difficulty are synthetically generated.

\subsubsection*{Step 0 - Parameters:}
The experimental framework takes two parameters for the generation of an artificial ground truth: $\alpha$, which controls the difficulty of the instances, and $r$, which is the number of possible class labels. There are three parameters that control the generation of the candidate sets: $\beta_0$, which controls the degree of hesitation of the annotators, $l$, which is the number of available annotators, and $s$, which is the maximum size of the candidate sets, where $s=1$ corresponds to full labeling and $s>1$ to candidate labeling. The number of instances is fixed to $n=500$ and the process of labeling is repeated $m=100$ times for each instance. The selection of these values aims to ensure stable results that are barely affected by randomness.

\subsubsection*{Step 1 - Ground truth generation:}

Given a specific parametrization, a domain is generated and represented by a Dirichlet distribution with hyperparameters $\alpha_c=\alpha$, for $c=1,...,r$. The Dirichlet distribution is used for sampling $n$ probability distributions over the class labels, which represent the difficulty of the $n$ instances. For each instance, the difficulty distribution is denoted by $d(c)$, and the ground truth label corresponds to $c^*=\arg\!\max_c d(c)$. As $d(c)$ becomes more uniform, the difficulty of the instance increases. On the contrary, as $d(c^*)\to 1$, the difficulty of the instance decreases. Due to the properties of the Dirichlet distribution, on average, a high value of $\alpha$ produces uniform distributions and, in consequence, difficult instances. On the contrary, a low $\alpha$ value results in easy instances on average. After repeating this step for the $n$ instances, the set of ground truth labels $\{c_i^*\}_{i=1}^n$ is obtained. This is used in Step 4 for estimating the error.

\subsubsection*{Step 2 - Annotator simulation:}
The model of annotator used in this work consists of two parts: i) the behavior of the annotator, which brings together their knowledge and the way they handle it, and ii) the labeling process given their behavior. The behavior is represented with a probability distribution over the class labels and the labeling process is simulated by means of a random sampling.

For each instance, taking into account its difficulty, $d(c)$, a set of $l$ annotators is generated. As noted before, the behavior of an annotator regarding an instance is modeled by a distribution $b(c)$, where $b(c)$ represents the preference of the annotator towards the class label $c$. The distribution $b(c)$ is obtained by sampling a Dirichlet distribution with parameters $\beta_0 \cdot d(c)$, for $c=1,...,r$. Thus, the behavior of an annotator depends on the difficulty of the instance $d(c)$ and the parameter $\beta_0$. The parameter $\beta_0$ can be seen as the average degree of hesitation of the annotators. For instance, as $\beta_0$ tends to~$0$, the behavior distribution will concentrate on a single label, that is, $b(c)\to 1$ for some class label $c$. Note that this does not mean that the annotator is right but he/she just has a very low degree of hesitation. On the other hand, as $\beta_0\to \infty$, the behavior $b(c)$ becomes more similar to the difficulty $d(c)$. Reasonably, in no scenario does the behavior of an annotator improve the instance difficulty.

Once the behavior distributions of the annotators $\{b_j(c)\}_{j=1}^l$ are fixed, the full and candidate labeling of the instance are simulated. For this purpose, a random sampling (with replacement) of size $s$ of the distributions, $b_j(c)$, is performed for each annotator ($j\in\{1,\ldots,l\}$). The parameter $s$ controls the flexibility with which the annotators handle their knowledge to produce a candidate set ---e.g., $s=1$ corresponds to full labeling (no flexibility) while higher values of $s$ correspond to candidate labeling (greater flexibility). Note that when the value of $s$ increases, the probability that the correct class appears in a candidate set becomes higher, but the probability of selecting other classes also grows. All the distinct class labels that are sampled from $b_j(c)$ form the candidate set $S_j$ of annotator $j$ ($j\in\{1,\ldots,l\}$). Thus, the size of the candidate sets is upper-bounded by $s$.

%In short, annotators are obtained by sampling Dirichlet distributions with different parameters ($\alpha$ and $\beta_0$) that characterize the population of annotators. So, every instance is labeled by a different set of $l$ annotators selected from that specific population. 

\subsubsection*{Step 3 - Voting:}
Given the set of candidate sets $L$ provided for an instance, an aggregated label is produced by means of the full (Eq. \ref{eq:voting}) or candidate voting (Eq. \ref{eq:cVoting}): $\hat{c}=\nu(L)$. By repeating the labeling and voting processes $m$ times, $m$ estimated ground truth values are obtained, $\{\hat{c}^k\}_{j=1}^m$. 
Finally, by repeating Steps 1 to 3 $n$ times, the set of multisets $\left\{\{\hat{c}_i^k\}_{k=1}^m\right\}_{i=1}^n$ is obtained.

\subsubsection*{Step 4 - Error estimation:}
%The goal of the experimentation is to estimate and analyze the error of the voting $\nu$ for different values of $s$ and in different settings.
The goal of the experiments is to estimate and analyze in different settings the error of the voting $\nu$ for different values.

Let us define $fr_i(c) = \frac{1}{m}\sum_{k=1}^m{\mathds{1}(\hat{c}_i^k=c)}$ as the \emph{frequency} of the label $c$ among the estimated ground truth values of instance~$i$. The error of $\nu$ is estimated as follows: 
\begin{equation}
\label{eq:error}
\hat{\epsilon}(\nu)= \frac{1}{n} \sum_{i=1}^n 1-(fr_i(c_i^*))\nonumber
\end{equation}

Additionally, following the methodology of \cite{kohavi97}, the estimated error $\hat{\epsilon}(\nu)$ can be decomposed into the bias and variance terms:
% \begin{equation}
% \label{eq:decomp}
% \hat{\epsilon}(\nu)= \delta^2 + \sigma^2\nonumber
% \end{equation}
% where $$\delta^2 = \frac{1}{2 n} \cdot \sum_{i=1}^n \left((1-fr_i(c_i^*))^2+\sum_{c\neq c_i^*}^r ( fr_i(c))^2\right)$$ corresponds to the squared bias and the variance term is $$\sigma^2= \frac{1}{2 n}\cdot\sum_{i=1}^n (1-\sum_{c=1}^r (fr_i(c))^2).$$
%
\begin{eqnarray}
\label{eq:decomp}
\nonumber \hat{\epsilon}(\nu)= & \underbrace{\frac{1}{2 n} \cdot \sum_{i=1}^n \left((1-fr_i(c_i^*))^2+\sum_{c\neq c_i^*}^r ( fr_i(c))^2\right)}_{\text{Squared bias}} \\
 & + \underbrace{\frac{1}{2 n}\cdot\sum_{i=1}^n (1-\sum_{c=1}^r (fr_i(c))^2)}_{\text{Variance}}
\end{eqnarray}

\subsection{Data generation: Case~R}

In Case~R, real-world datasets are used to obtain the difficulty distributions and the ground truth. Case~A and Case~R only differ in the number of parameters used (Step $0$) and in the way the ground truth and the difficulty distributions are generated (Step $1$). Three parameters are used in Case~R: $\beta_0$ (hesitation level of annotators), $l$ (number of annotators) and $s$ (flexibility). The number of instances $n$ is given by the used dataset. As in Case~A, in order to ensure stable results, the labeling process is repeated $m=100$ times for each instance. 

Step $1$ for Case~R is as follows: Real data is used in order to get more realistic labels. The difficulties are obtained through Random Forest classifiers \cite{breiman01}. A fair validation is used to obtain class probabilities $p(c|i)$ for each instance $i$ from a Random Forest not trained with $x$. In order to avoid zero probabilities, we add a smoothing vector with values $\frac{1}{r}$ to the class probabilities. The resulting sum is normalized to obtain the difficulty distribution: $d(c)= \frac{r*p(c|i)+1}{2r}$. Note that, in this case, as opposed to the Case~A, when an instance is mistakenly classified by the random forest, we have $c^*\neq \argmax_c d(c)$. The error achieved in a dataset by the Random Forest sets a lower bound for the error values that %the different voting schemes will achieve.
a voting scheme can reach.

\subsubsection*{Checking the hypotheses:} 
By means of the proposed experimental framework, the soundness of the hypotheses presented in Section \ref{sec:candidatecrowd} can be checked, analyzing the effects of different parameters in the estimated errors of full and candidate voting:
  \begin{itemize}
      \item \textbf{H1}: This hypothesis can be checked by using different values of the parameter $l$.

      \item \textbf{H2}: By varying the value of parameter $\alpha$, and in order to observe its effect on the performance of the different voting schemes, easier or more difficult instances can be generated, 

      \item \textbf{H3}: Through the value of the parameter $r$, the influence of the number of possible class labels can be tested. 

      \item \textbf{H4}: The parameter $\beta_0$ allows us to control the degree of hesitation of the annotators to check this hypothesis.

\end{itemize}

\subsection{Case A: Empirical results with artificial data}
\label{sec:artresults}

\begin{figure*}[!ht]
\centering
\subfloat[][$r=32,\ \alpha=0.5,\ \beta_0=4$]{\label{fig:errr32hyp05b04}
\includegraphics[width=0.31\textwidth]{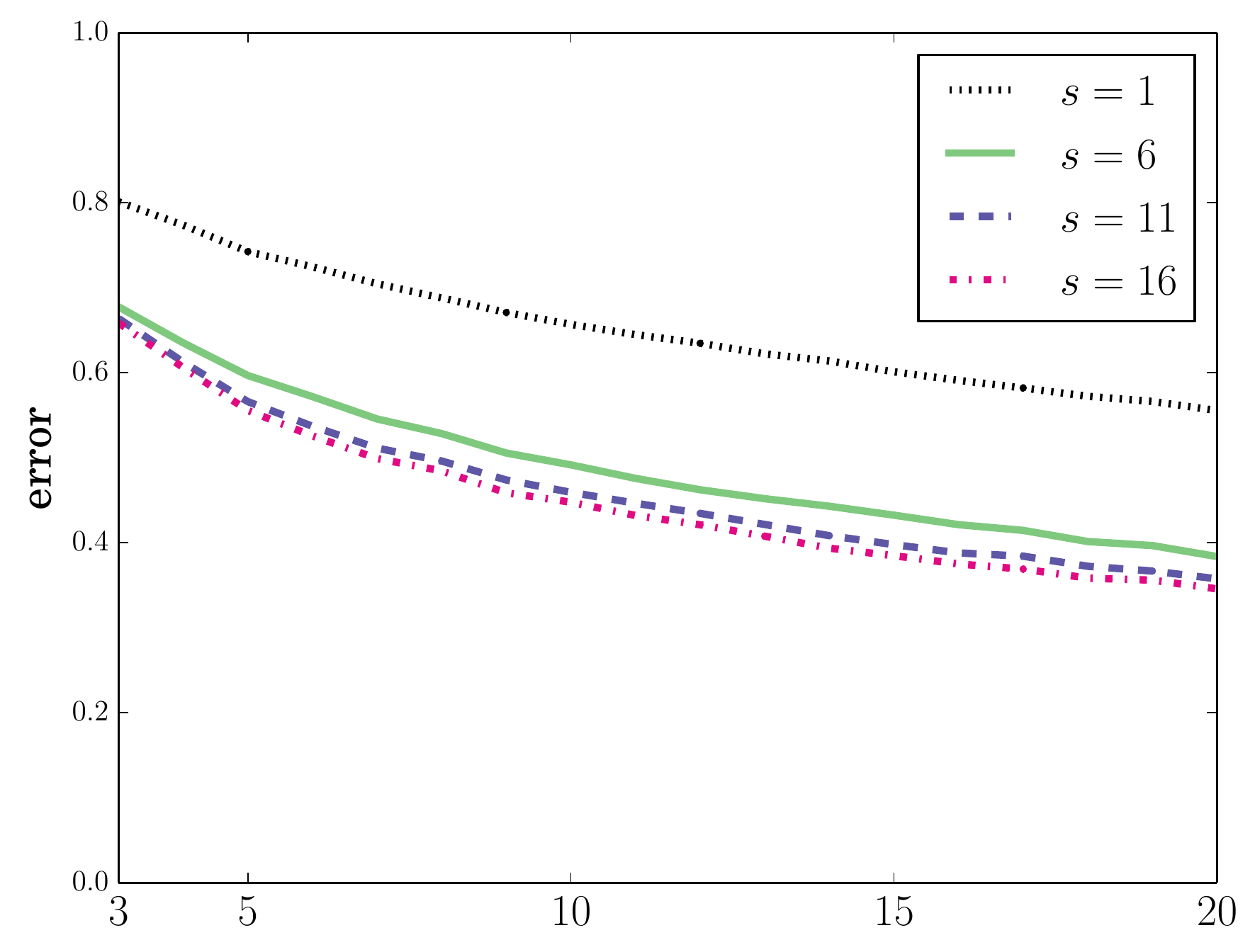}}%\hspace{0.1\textwidth}
%\subfloat[][$r=32,\ \alpha=0.5,\ \beta_0=0.25$]{\label{fig:errr32hyp05b0025}
%\includegraphics[width=0.23\textwidth]{01lossAllnotweightedr32hyp05b0025.eps}}%\\
\subfloat[][$r=32,\ \alpha=10,\ \beta_0=4$]{\label{fig:errr32hyp10b04}
\includegraphics[width=0.31\textwidth]{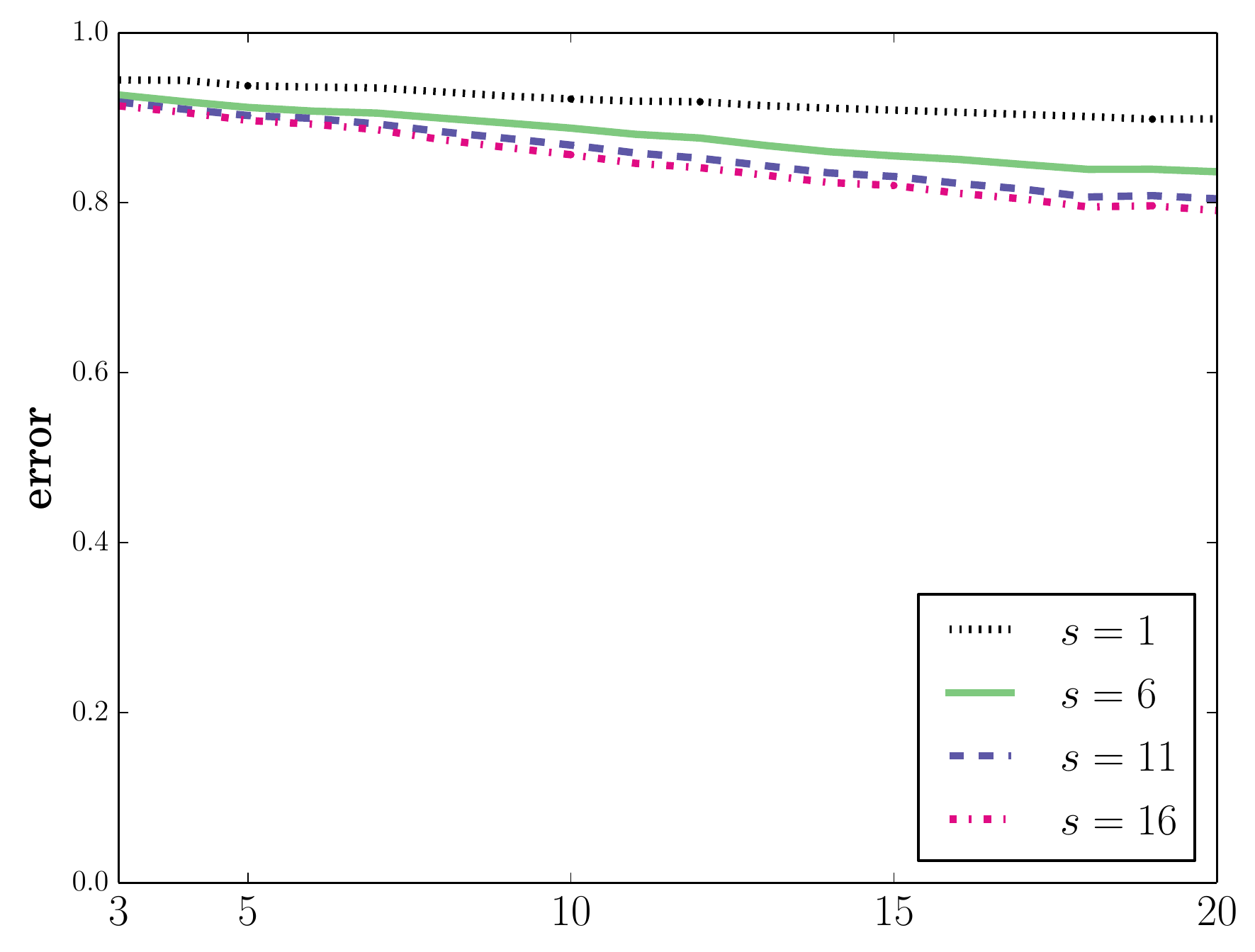}}%\hspace{0.1\textwidth}
\subfloat[][$r=8,\ \alpha=0.5,\ \beta_0=4$]{\label{fig:errr8hyp05b04}
\includegraphics[width=0.31\textwidth]{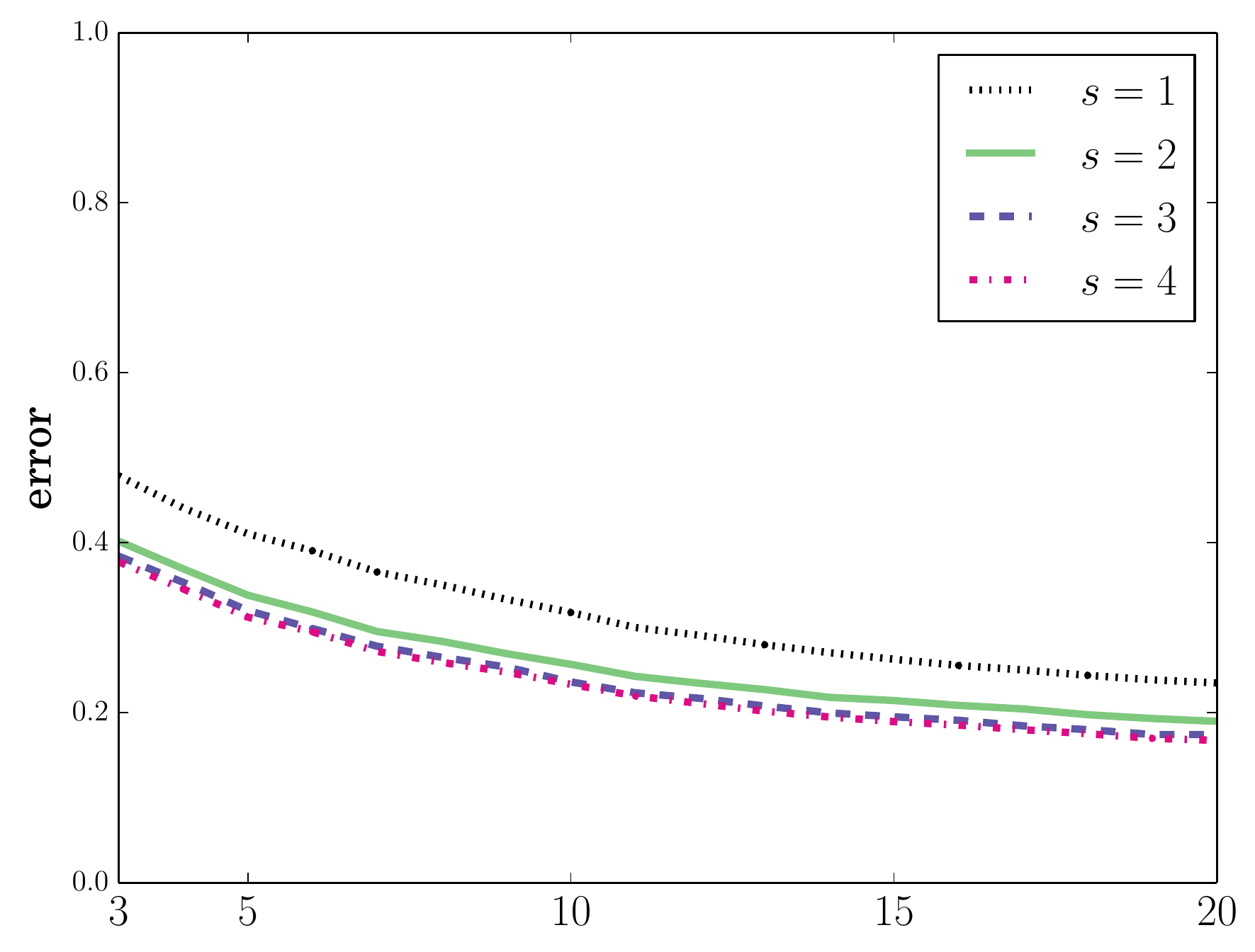}}
\caption{Graphical description of the error (Eq. \ref{eq:error}) obtained by annotation simulated with different values of $r$ and $\alpha$. Error curves for different values of $s$ are shown in each figure.}
\label{fig:totalerr}
\end{figure*}
\begin{figure*}[!ht]
\centering
\subfloat[][$r=32,\ \alpha=0.5,\ l=8$]{\label{fig:trader32l8alpha05}
\includegraphics[width=0.29\textwidth]{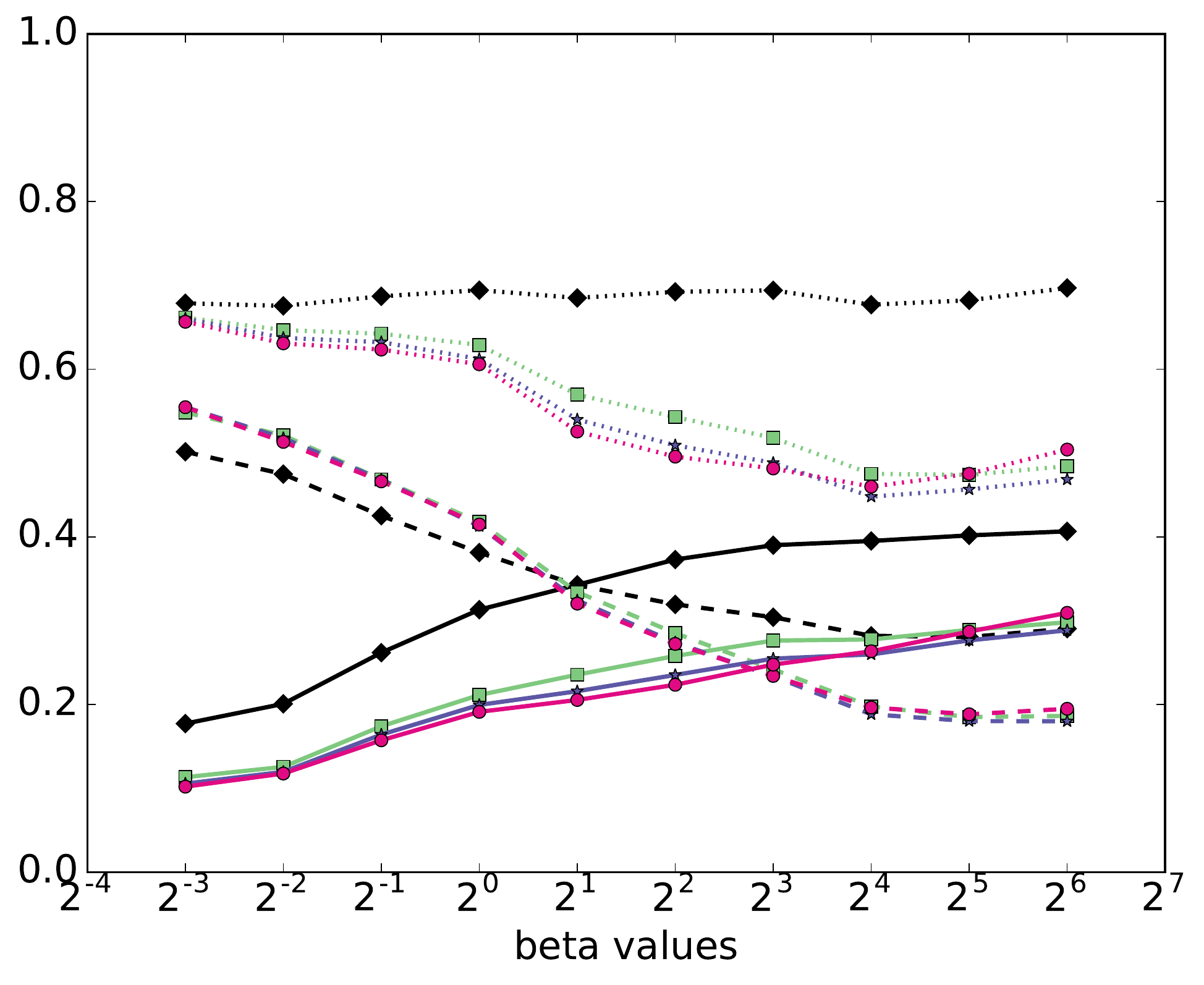}}%\hspace{0.1\textwidth}
\subfloat[][$r=32,\ \alpha=2,\ l=8$]{\label{fig:trader32l8alpha2}
\includegraphics[width=0.29\textwidth]{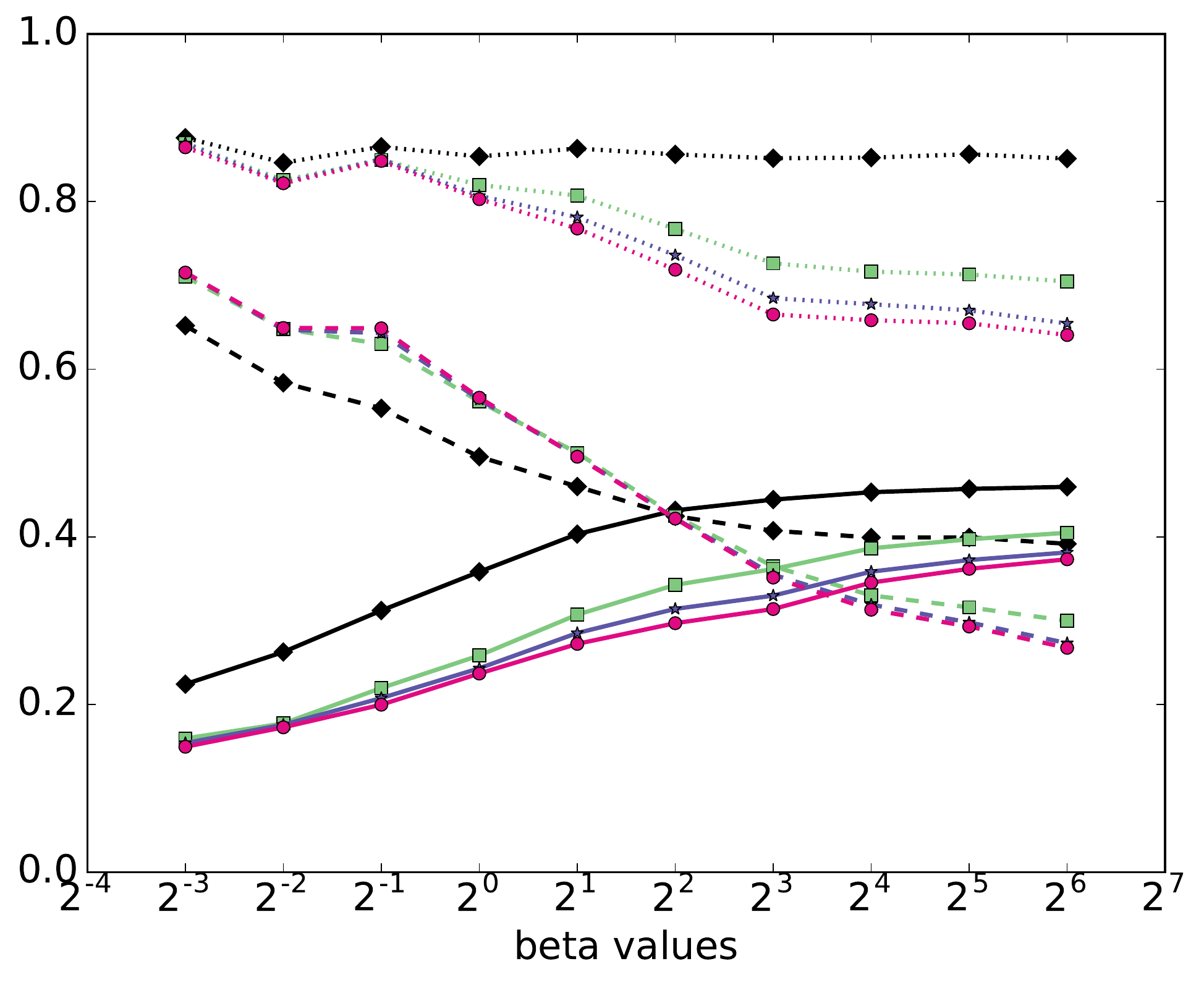}}
\subfloat[][$r=32,\ \alpha=10,\ l=8$]{\label{fig:trader32l8alpha10}
\includegraphics[width=0.365\textwidth]{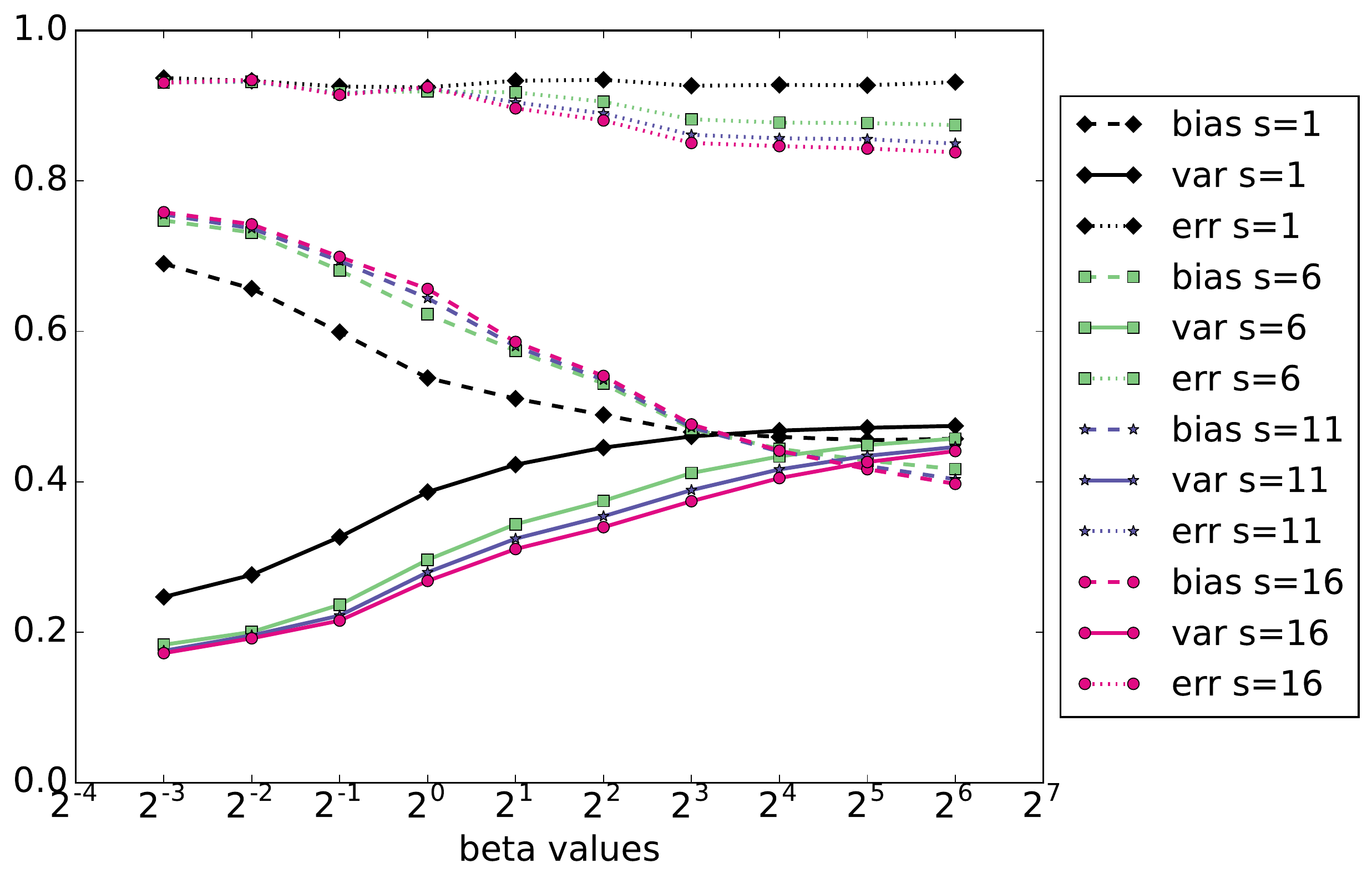}}%\hspace{0.1\textwidth}

\caption{Graphical description of the decomposition of the error obtained by annotation simulated with different values of $\alpha$ and $s$. Curves representing the squared bias, the variance and the total error are displayed in each figure.}
\label{fig:tradeoff}
\end{figure*}

Using the framework proposed for Case~A, different domains and annotators have been generated by setting different values for the parameters $r$, $\alpha$, $\beta_0$, $s$ and $l$.  %Experiments were carried out with larger values and the results were similar.

Experimental results are graphically summarized in Figs.~\ref{fig:totalerr} (total error) and \ref{fig:tradeoff} (bias/variance trade-off). Fig.~\ref{fig:totalerr} shows the results for the combination of difficult ($\alpha=10$) and easy ($\alpha=0.5$) instances, with different numbers of class labels $r\in \{8,32\}$. In each plot, the parameters $\alpha$ and $r$ have been fixed at extreme values in order to observe the differences that they cause in the results. Each plot shows error curves obtained with different values of $s<\frac{r}{2}$ and an increasing number of annotators, $3<l<20$. For all three plots in Fig.~\ref{fig:totalerr}, an intermediate value for the parameter that controls the hesitation of the annotators, $\beta_0=4$, has been selected in order to avoid its influence on the displayed results. In Fig.~\ref{fig:tradeoff}, the reader can observe the effect of increasing the hesitation of the labelers (parameter $\beta_0$) in the bias/variance trade-off, and analyze the source of the error in each scenario. In that figure, results with different difficulty degrees ($\alpha\in\{0.5,2,10\}$), $r=32$ possible class labels and $l=8$ annotators are shown. Additional experiments with different settings have been carried out and the corresponding figures are available in the supplementary material\footnote{\url{https://github.com/isg-ehu/iker.benaran}}.

The error becomes lower as $l$ increases in both the full labeling and the candidate labeling scenarios (see Fig.~\ref{fig:totalerr}), i.e., more labelers provide more knowledge. Also, with both labeling approaches, better performance is observed in domains with a lower number of possible class labels ($r=8$ in Fig.~\ref{fig:errr32hyp05b04} as opposed to $r=32$ in Fig.~\ref{fig:errr8hyp05b04}) and in easier domains (see $\alpha=0.5$ in Figs.~\ref{fig:errr32hyp05b04} and \ref{fig:trader32l8alpha05} as opposed to $\alpha=10$ in Figs.~\ref{fig:errr32hyp10b04} and \ref{fig:trader32l8alpha10}).

According to these results, the full voting ($s=1$) has consistently a poorer performance than the candidate voting ($s>1$), which would support our hypothesis \textbf{H1}. For example, in the scenario displayed in Fig.~\ref{fig:errr8hyp05b04}, full labeling needs at least $10$ annotators to achieve the same error as even the least flexible candidate labeling approach ($s=6$ in Figs.~ \ref{fig:errr32hyp05b04} and \ref{fig:errr32hyp10b04}, $s=2$ in Fig.~\ref{fig:errr8hyp05b04}), and the error obtained by candidate labeling with $l=5$ annotators is only achieved by full labeling with at least $l=10$ annotators. Moreover, in Fig.~\ref{fig:errr32hyp05b04}, the error shown by candidate labeling ($s\in\{6,11,16\}$) with $l=5$ annotators is only observed with $l=15$ annotators when using full labeling. In Fig.~\ref{fig:errr32hyp10b04}, $l=20$ annotators with full labeling are required to perform as well as candidate labeling ($s\in\{6,11,16\}$) with $l=5$ labelers.

The more difficult the domain (i.e., the higher the parameter $\alpha$ is), the larger is number of annotators required by full labeling in order to achieve a similar or lower error than candidate labeling (Fig.~\ref{fig:errr32hyp10b04} vs. Fig.~\ref{fig:errr32hyp05b04}). In other words, candidate labeling seems to take more advantage of the increasing number of annotators in difficult instances than in easy instances. Moreover, in a difficult domain, full labeling barely profits from the increasing number of annotators (see Fig.~\ref{fig:errr32hyp10b04}). For example, to achieve a similar error as $4$ annotators with candidate labeling, in a difficult domain ($\alpha=10$, Fig.~\ref{fig:errr32hyp10b04}), $20$ annotators with full labeling are required, while less than $15$ annotators are sufficient in an easy domain ($\alpha=0.5$, Fig.~\ref{fig:errr32hyp05b04}). These last facts provide evidence for \textbf{H2}.

With a high number of possible class labels ($r=32$), the difference between the error curves of full and candidate labeling (in all its different levels) becomes bigger, as hypothesized in \textbf{H3} (Fig. \ref{fig:errr32hyp05b04} vs. Fig. \ref{fig:errr8hyp05b04}). Many other experimental scenarios that show the same tendency and support both \textbf{H2} and \textbf{H3} are compiled in the supplementary material.

As stated above, the effect of the parameter $\beta_0$ on the error and the bias/variance trade-off can be seen in Fig.~\ref{fig:tradeoff}. When the hesitation ($\beta_0$) of the labelers increases, the error curve of full labeling remains quite similar, while the error curves of different levels of candidate labeling decrease noticeably. In other words, similar results are obtained in full labeling with hesitant and obstinate labelers, while candidate labeling takes advantage of the hesitant labelers. Thus, in this experimental setting, the hypothesis \textbf{H4} holds.

A rise in the error curve of candidate labeling can be observed when there is an extremely high hesitation degree (large $\beta_0$) along with easy instances (small $\alpha$). This can be clearly observed in Fig.~\ref{fig:trader32l8alpha05}. In easy domains, usually a few of the class labels have a much higher probability than the rest in the difficulty distribution $d(c)$. That can be interpreted as a dependence relation between those highly probable class labels. Moreover, due to the high value of $\beta_0$, that dependence also appears in the behavior distributions $b(c)$. In that scenario, for sufficiently high values of $s$, all the candidate sets contain these highly probable class labels. Consequently, all these labels get the same number of votes, and the candidate voting results in a draw. As draws are solved randomly, candidate voting may be wrong even if the correct class label was selected in all the candidate sets.

As for the bias-variance trade-off, on the one hand, low values of $\beta_0$ cause unbalanced behavior distributions $b(c)$. On the other hand, high values of $\beta_0$ lead to behavior distributions that are similar to the previously generated difficulty distribution $d(c)$, which can be either unbalanced or uniform (depending on the value of $\alpha$). In the scenarios where $b(c)$ is unbalanced, similar results tend to occur when performing the $m=100$ repetitions of the sampling process. Thus, the mistaken guesses are concentrated in few class labels, so the error is mostly caused by bias. That can be seen in any plot displayed in Fig.~\ref{fig:tradeoff}. The effect of high hesitation degrees (large $\beta_0$) combined with easy instances (small $\alpha$) can be observed particularly in Fig.~\ref{fig:trader32l8alpha05}. In scenarios with uniform $b(c)$ distributions, different results are reached when repeating the sampling process, so variance becomes the main source of the error. 

\subsection{Case R: Empirical results with real supervised data}

\begin{table}
\centering\scriptsize
\begin{tabular}{ c c c c c }

\hline
\textbf{Dataset} & \textbf{\# inst.} & \textbf{\# attributes} & \textbf{\# classes} & \textbf{RF error} \\
    \hline
    %\hline
    arrhythmia & $452$ & $279$ & $13$ & $0.334$ \\
    vowel & $990$ & $10$ & $11$ & $0.353$ \\
    segment & $2310$ & $19$ & $7$ & $0.026$ \\
    letter & $20000$ & $16$ & $26$ & $0.059$ \\
    mnist & $60000$ & $780$ & $10$ & $0.056$ \\
    \hline

\end{tabular}
\label{tab:datasets}
\caption{Features of the datasets used for experiments: Number of instances, number of attributes, number of classes and error achieved with the Random Forest classifier.}
\end{table}

In this set of experiments, five real-world datasets from the UCI repository~\cite{frank10} are used within the framework described in Section \ref{sec:data} for Case~R. For each dataset, two different scenarios ($l\in\{4,8\}$ annotators) are set.

%%%%%%%%%%%%%%%%%%%%%%%%%%%%%%%%%%%%%%%%%%%%%%%%%%%%%%%%%%%%

\begin{table}[t]
\centering
\tiny
%\footnotesize
\begin{tabular}{ cc|c|c|c|c|c|c| }
\cline{3-8}
 &  & \multicolumn{3}{|c|}{$l=4$} & \multicolumn{3}{|c|}{$l=8$}  \\
\cline{3-8}
Datasets & $s$ & $\beta_0=1$ & $\beta_0=4$ & $\beta_0=16$ &  $\beta_0=1$ & $\beta_0=4$ & $\beta_0=16$ \\
\hline

\multicolumn{1}{ |c }{\multirow{2}{*}{arrhythmia}} & \multicolumn{1}{ |c| }{$1$} & $0.622$ & $0.619$ & $0.620$ & $0.514$ & $0.514$ & $0.512$ \\ \cline{2-8}
 \multicolumn{1}{ |c  }{\multirow{2}{*}{RF err: 0.334}}
  & \multicolumn{1}{ |c| }{$4$} & $0.551$ & $0.496$ & $0.466$ & $0.456$ & $0.419$ & $0.403$ \\ \cline{2-8}
  \multicolumn{1}{ |c  }{}
   & \multicolumn{1}{ |c| }{$7$} & $0.538$ & $0.472$ & $\bd{0.454}$ & $0.444$ & $0.408$ & $\bd{0.395}$ \\ \hline

\multicolumn{1}{ |c }{\multirow{2}{*}{vowel}} & \multicolumn{1}{ |c| }{$1$} & $0.638$ & $0.639$ & $0.640$ & $0.542$ & $0.543$ & $0.543$ \\ \cline{2-8}
\multicolumn{1}{ |c  }{\multirow{2}{*}{RF err: 0.353}}
  & \multicolumn{1}{ |c| }{$3$} & $0.585$ & $0.543$ & $0.517$ & $0.492$ & $0.457$ & $0.440$ \\ \cline{2-8}
  \multicolumn{1}{ |c  }{}
   & \multicolumn{1}{ |c| }{$5$} & $0.572$ & $0.516$ & $\bd{0.493}$ & $0.479$ & $0.440$ & $\bd{0.421}$ \\ \hline

\multicolumn{1}{ |c }{\multirow{2}{*}{segment}} & \multicolumn{1}{ |c| }{$1$} & $0.266$ & $0.267$ & $0.264$ & $0.125$ & $0.124$ & $0.124$ \\ \cline{2-8}
\multicolumn{1}{ |c  }{\multirow{2}{*}{RF err: 0.026}}
  & \multicolumn{1}{ |c| }{$2$} & $0.202$ & $0.160$ & $0.135$ & $0.084$ & $0.062$ & $0.054$ \\ \cline{2-8}
  \multicolumn{1}{ |c  }{}
   & \multicolumn{1}{ |c| }{$3$} & $0.183$ & $0.128$ & $\bd{0.099}$ & $0.073$ & $0.051$ & $\bd{0.043}$ \\ \hline

\multicolumn{1}{ |c }{\multirow{2}{*}{letter}} & \multicolumn{1}{ |c| }{$1$} & $0.376$ & $0.377$ & $0.377$ & $0.191$ & $0.192$ & $0.191$ \\ \cline{2-8}
\multicolumn{1}{ |c  }{\multirow{2}{*}{RF err: 0.059}}
  & \multicolumn{1}{ |c| }{$3$} & $0.235$ & $0.141$ & $\bd{0.119}$ & $0.117$ & $0.091$ & $\bd{0.083}$ \\ \cline{2-8}
  \multicolumn{1}{ |c  }{}
   & \multicolumn{1}{ |c| }{$5$} & $0.205$ & $0.128$ & $0.109$ & $0.086$ & $0.085$ & $0.085$ \\ \hline

\multicolumn{1}{ |c }{\multirow{2}{*}{mnist}} & \multicolumn{1}{ |c| }{$1$} & $0.369$ & $0.369$ & $0.368$ & $0.216$ & $0.215$ & $0.215$ \\ \cline{2-8}
\multicolumn{1}{ |c  }{\multirow{2}{*}{RF err: 0.056}}
  & \multicolumn{1}{ |c| }{$3$} & $0.280$ & $0.218$ & $0.181$ & $0.152$ & $0.119$ & $0.105$ \\ \cline{2-8}
  \multicolumn{1}{ |c  }{}
   & \multicolumn{1}{ |c| }{$5$} & $0.258$ & $0.184$ & $\bd{0.156}$ & $0.138$ & $0.105$ & $\bd{0.093}$ \\ \hline

\end{tabular}
\label{tab:results}
\caption{Error rates for simulations using different datasets and different values of the parameters $\beta_0$, $l$ and $s$. The lowest error obtained in each scenario is in bold. \emph{RF err} refers to the classification error shown by a Random Forest trained in the given dataset.}
\end{table}

Experimental results can be observed in Table \ref{tab:results}. For every dataset, the scenarios range from obstinate ($\beta_0=1$) to hesitant ($\beta_0=16$) annotators and varying flexibilities ($s$). The error reached with different values of the parameters $s$ (flexibility) and $\beta_0$ (level of hesitation) are compared in each scenario (fixed dataset and number of annotators), and the lowest error obtained in each scenario is highlighted in bold. For every dataset, the error obtained with the Random Forest classifier is shown as a lower bound for the error achieved through both full and candidate labeling. As could be expected, the datasets that obtain the lowest errors with the Random Forest classifier also obtain the lowest error values through the simulated labeling and candidate voting.

As in Case A, the error also decreases when there are more annotators available (see $l=4$ against $l=8$). In all the cases with $\beta_0>1$ --- except for $\beta_0=4$ with dataset \emph{segment} --- the error of candidate labeling ($s>1$) is equal to or lower than that of the full labeling ($s=1$), which supports our hypothesis \textbf{H1}. Hypothesis \textbf{H2} cannot be checked within this framework, as the difficulties are determined by the Random Forest classifier and not by a fixed parameter, as $\alpha$ in the Case~A. Hypothesis \textbf{H3} cannot be contrasted because the number of classes $r$ cannot be isolated from the rest of elements %(especially the choice of the dataset).
(especially, dataset selection).

Experimental scenarios with $s$ values near $\frac{r}{2}$ show the best results. As the value of $s$ increases, the error becomes lower in almost every case --- except for dataset \emph{letter} with $\beta_0=16$ --- both with $l=4$ and $l=8$. This fact suggests that, when a labeler is more flexible, more information can be extracted. With the \emph{letter} dataset, the number of possible class labels is $r=26$. Sampling a behavior distribution either $7$ or $13$ times may bring similar results, with a little overfitting when there are hesitant annotators. Note that candidate labeling ($s>1$) always obtains a lower error than full labeling ($s=1$), reaching error values similar to those obtained by the Random Forest classifier on each dataset, i.e., close to the lower bound. %Recall that the Random Forest classifier sets the difficulties of the instances, so the error value obtained with that classifier is the minimum that can be reached by aggregating the candidate sets given by the annotators.

Similarly to Case~A, candidate labeling ($s>1$) profits from hesitant labelers (larger~$\beta_0$): The larger the value of~$\beta_0$, the lower the error. On the contrary, the error reached with full labeling ($s=1$) barely changes from obstinate ($\beta_0=1$) to hesitant ($\beta_0=16$) annotators. Thus, as the hesitation level increases, candidate labeling outperforms the full labeling approach, which would support our hypothesis~\textbf{H4}.

To sum up, these tests pose empirical evidence of the presented four hypothesis and show that candidate labeling gathers more information of supervision than full labeling in crowdsourced annotations. Candidate labeling is especially useful with a low number of workers, with difficult instances, with hesitant workers and/or with a large number of possible labels.

\section{Conclusion and future work}
\label{sec:conclusions}
In this work, candidate labeling is proposed as an alternative to the traditional full labeling for crowd learning. The workers are given the possibility of choosing various labels for each instance instead of just one. Intuitively, this simple mechanism can help extract more knowledge than the full labeling from a set of annotators in learning from crowds problems.

Throughout an experimental framework with artificial and real-world data, empirical evidence suggests that the use of candidate labeling could be particularly profitable compared to full labeling when (i) the number of available annotators is low, (ii) the difficulty of the instance is high, (iii) the number of possible class labels is high, or when (iv) annotators are hesitant.

The results obtained with candidate labeling are promising. For future work, it would be interesting to collect candidate sets from a real crowd for a real domain. In order to verify the hypotheses in a fair experimental setting, an experiment has been prepared\footnote{\label{foot:whichdog}http://bit.ly/which-dog}. In this experiment, workers are asked to label images from the Stanford Dogs dataset \cite{khosla11} with the breed of the dogs, and annotators are permitted to choose as many labels as they want from a list of possible breeds. From the analysis of the collected labels, we expect to provide new insights on the usefulness of weak supervision for crowdsourced labeling. Voting schemes that take into account the reliability of annotators could also be conceived, as well as learning techniques for datasets labeled with candidate labeling. Those techniques might include a model for the expertise of annotators.

It could be interesting to study whether candidate labeling implies a larger effort for a labeler than the traditional full labeling. On the one hand, it may take more time to select various labels instead of just one but, on the other hand, an annotator could find it easier to provide a candidate set rather than to try to find out which one of them is the correct one. As mentioned in the introduction, partial evidence has already been provided \cite{banerjee17} about the intuition that annotators work faster using candidate labeling than full labeling.

\emph{This paper is under consideration at Pattern Recognition Letters.}

% if have a single appendix:
%\appendix[Proof of the Zonklar Equations]
% or
%\appendix  % for no appendix heading
% do not use \section anymore after \appendix, only \section*
% is possibly needed

%\appendices
%\section{Proof of the First Zonklar Equation}

% you can choose not to have a title for an appendix
% if you want by leaving the argument blank
%\section{}

% use section* for acknowledgment
\ifCLASSOPTIONcompsoc
  % The Computer Society usually uses the plural form
  \section*{Acknowledgments}
\else
  % regular IEEE prefers the singular form
  \section*{Acknowledgment}
\fi

% \section{ Acknowledgments}

J. Hern\'andez-Gonz\'alez is partially supported by the Basque Government (IT609-13, Elkartek BID3A), the Spanish Ministry of Economy and Competitiveness MINECO (TIN2016-78365-R) and the University-Society Project 15/19 (Basque Government and UPV/EHU). I. Be\~naran-Mu\~noz and A. P\'erez are both supported by the Spanish Ministry of Economy and Competitiveness MINECO through BCAM Severo Ochoa excellence accreditation SEV-2013-0323 and the project TIN2017-82626-R funded by (AEI/FEDER, UE). I. Be\~naran-Mu\~noz is also supported by the grant BES-2016-078095. A. P\'erez is also supported by the Basque Government through the BERC 2014-2017 and the ELKARTEK programs, and by the MINECO through BCAM Severo Ochoa excellence accreditation SVP-2014-068574.

% references section

\begin{IEEEbiography}[{\includegraphics[width=1in,height=1.25in,clip,keepaspectratio]{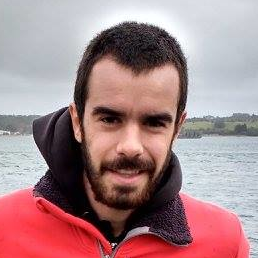}}]{Iker Be\~naran-Mu\~noz} obtained his Bachelor's Degree in Mathematics at the University of the Basque Country (UPV/EHU) in July 2015. He finished a Master's Degree in Computational Engineering and Intelligent Systems at the same university in September 2016. He is currently a Ph.D. Student at BCAM, in the area of Data Science.

\end{IEEEbiography}

\begin{IEEEbiography}[{\includegraphics[width=1in,height=1.25in,clip,keepaspectratio]{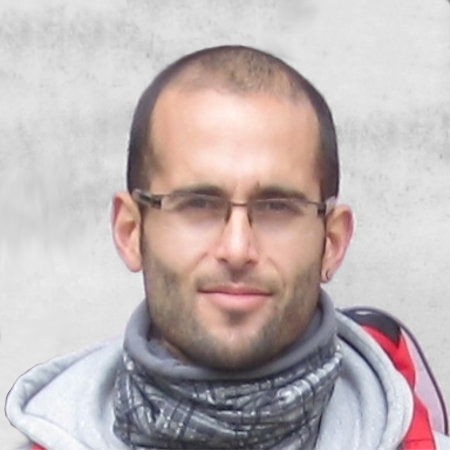}}]{Jer\'onimo Hern\'andez-Gonz\'alez} received his Ph.D. in computer science from the \emph{University of the Basque Country} in 2015. He is a post-doc researcher in the \emph{University of the Basque Country} and member of the \emph{Intelligent Systems Group}. His main research interests are (weakly) supervised learning problems, Bayesian networks and computational biology. He has 11 publications in refereed journals and conferences. 
\end{IEEEbiography}

\begin{IEEEbiography}[{\includegraphics[width=1in,height=1.25in,clip,keepaspectratio]{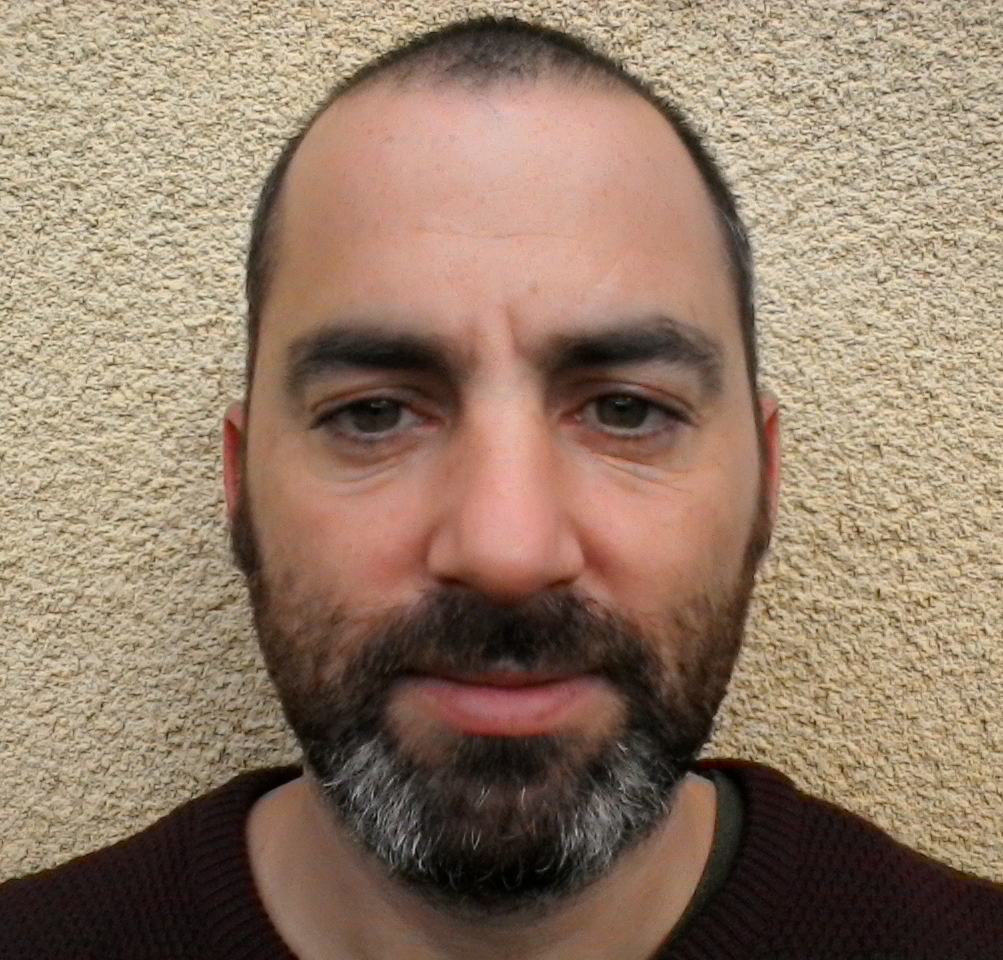}}]{Aritz P\'erez} received his Ph.D. degree in 2010 from the the University of Basque Country, department of Computer Science and Artificial Intelligence. Currently, he is a postdoctoral researcher at the Basque Center for Applied Mathematics. His current scientific interests includes supervised, unsupervised and weak classification, probabilistic graphical models, model selection and evaluation, time series and crowd learning.
\end{IEEEbiography}

%\vfill

\end{document}